\documentclass[authoryear,preprint, review, 12pt]{elsarticle}

\usepackage{graphicx}
\usepackage{amssymb}
\usepackage{subfigure}
\usepackage{multirow}
\usepackage{lineno}
\usepackage{amsmath}
\usepackage{natbib}
\UseRawInputEncoding
\setcitestyle{citesep={;}, aysep={,}}

\journal{Computers and Electronics in Agriculture}

\begin{document}

\begin{frontmatter}

\title{In-field high throughput grapevine phenotyping with a consumer-grade depth camera}

\author[label1]{Annalisa Milella}
\author[label1]{Roberto Marani}
\author[label1]{Antonio Petitti}
\author[label2]{Giulio Reina}

\address[label1]{Institute of Intelligent Industrial Technologies and Systems for Advanced Manufacturing, National Research Council, via G. Amendola 122/O, 70126 Bari, Italy. Corresponding author e-mail: annalisa.milella@stiima.cnr.it}
\address[label2]{University of Salento, Department of Engineering for Innovation, Via Arnesano, 73100 Lecce, Italy. }

\begin{abstract}
Plant phenotyping, that is, the quantitative assessment of plant traits including growth, morphology, physiology, and yield, is a critical aspect towards efficient and effective crop management. Currently, plant phenotyping is a manually intensive and time consuming process, which involves human operators making measurements in the field, based on visual estimates or using hand-held devices. In this work, methods for automated grapevine phenotyping are developed, aiming to canopy volume estimation and bunch detection and counting. It is demonstrated that both measurements can be effectively performed in the field using a consumer-grade depth camera mounted on-board an agricultural vehicle. First, a dense 3D map of the grapevine row, augmented with its color appearance, is generated, based on infrared stereo reconstruction. Then, different computational geometry methods are applied and evaluated for plant per plant volume estimation. The proposed methods are  validated through field tests performed in a commercial vineyard in Switzerland. It is shown that different automatic methods lead to different canopy volume estimates meaning that new standard methods and procedures need to be defined and established. Four deep learning frameworks, namely the AlexNet, the VGG16, the VGG19 and the GoogLeNet, are also implemented and compared to segment visual images acquired by the RGB-D sensor into multiple classes and recognize grape bunches. Field tests are presented showing that, despite the poor quality of the input images, the proposed methods are able to correctly detect fruits, with a maximum accuracy of 91.52\%, obtained by the VGG19 deep neural network.
\end{abstract}


\end{frontmatter}


\section{Introduction}\label{s1}
The precise knowledge of plant characteristics is essential for precision farming treatments, such as selective harvesting, precise spraying, fertilization and weeding, and for efficient and effective crop management in general. Traditionally, phenotypic evaluations are either done by visual inspections, hand-held devices or destructive sampling. These methods are labor intensive, they need to be done by skilled operators and may result in subjective estimates. \\In recent years, sensor-based approaches have been developed thanks to the availability of affordable sensors and electronic systems. Much effort has been especially devoted to the implementation of high-throughput phenotyping platforms (HTPPs), as a key tool to automatically assess the overall growth and development of the plant under controlled conditions, such as in environmentally controlled chambers or greenhouses. However, many of the agronomically relevant traits are best expressed under real environmental conditions, which are highly heterogeneous and difficult to reproduce. In particular, perennial crops like grapevines need phenotypic evaluations to be done directly in the field. Vineyards typically extend over large areas that contain thousands of single grapevines, each one characterized by a slightly different phenotype. Due to these reasons, recently, there has been an increasing interest in field-based HTPPs. In this respect, autonomous vehicles can play a central role towards the automation of data acquisition and processing for phenotyping of crop plants in the field, \citep{EM, REI_2017}. \\
In this work, a grapevine phenotyping platform is proposed that uses an agricultural vehicle equipped with a RGB-D sensor, i.e., a depth sensing device that works in association with a RGB camera by augmenting the conventional image with distance information in a per-pixel basis. The system is intended to be used to acquire visual and geometric 3D information to reconstruct the canopy of the plants for geometric measurements, such as plant volume estimates, and to detect and count grapevine clusters. Both measurements are important indicators of growth, health and yield potential of grapevines \citep{LIU}. Specifically, first, different computational geometry methods for plant modeling and volume estimation using 3D and color information acquired by the RGB-D sensor are proposed; then, a deep learning approach using visual images only is developed to segment the scene into multiple classes and in particular to detect grapevine clusters.\\ The proposed framework was validated through experimental tests carried out in a commercial vineyard in Switzerland, with an Intel RealSense R200 sensor mounted on-board a Niko caterpillar. Compared to traditional manual methods, the main advantage of the proposed approaches is that they allow for data collection and processing in the field during operations of an agricultural vehicle, using only a cost-effective 3D sensor. It is shown that, despite the low quality of the available sensor data, it is possible to extract useful information about the crop status in a completely automatic and non-invasive way, thus overcoming time and cost problems related to traditional man-made measurements.\\The rest of the paper is structured as follows. Section \ref{s2} reports related work. The data acquisition system and the proposed algorithms for volume estimation and bunch recognition are described in Section \ref{s3}. Experimental results are presented in Section \ref{s4}. Conclusions are drawn in Section \ref{s5}.

\section{Related Work}\label{s2}
Precision agriculture deals with the application of the right treatment, at the right place, at the right time \citep{LEGG}. To this end, the accurate knowledge of the crop characteristics, or phenotypic data, at sub-field level is crucial. Conventional approaches to estimate phenotypic traits are based on human labor using few random samples for visual or destructive inspection. These methods are time consuming, subjective and prone to human error, leading to what has been defined the phenotypic bottleneck \citep{FUR}. The lack of phenotypic information limits the possibility to assess quantitative traits, such as yield, growth and adaption to abiotic or biotic factors. To overcome this issue, high-throughput phenotyping platforms (HTPPs) using different sensor devices, either fixed or mounted on-board mobile vehicles are attracting much interest \citep{REI_2016}. Research efforts have been especially devoted to build up such platforms under controlled environments \citep{CLARK, HART}. Although these systems enable a precise non-invasive plant assessment throughout the plant life cycle under controlled conditions, they do not take into account the interactions between the plants and the environment. In contrast, perennial crops like grapevines need to be evaluated directly in the field. First steps towards a high-throughput phenotyping pipeline in grapevines have been introduced by \cite{HER} that developed a Prototype Image Acquisition System (PIAS) for semi-automated capturing of geo-referenced images and a semi-automated image analysis tool to phenotype berry size. The use of 2D machine vision techniques has been also proposed in several works to detect grape bunches \citep{FONT,DIA}, and to measure berry size or number \citep{ROS,NUS}.\\While producing good results, 2D approaches are affected by occlusions. Furthermore, they do not allow to catch important phenotypic parameters such as 3D plant geometry. To this end, solutions employing stereovision or multi-view-stereo (MVS) systems have been proposed \citep{REI_2012}. Few of them have been developed in the context of viticulture. \cite{DEY} use a MVS algorithm on images manually collected with a consumer camera to reconstruct grapevine rows, and developed a Support Vector Machine (SVM) classifier to segment the point cloud into grape bunches, leaves and branches. A MVS approach is also developed by \cite{HH} to reconstruct point clouds of slightly defoliated grapevines from images acquired by a hand-held camera at daytime. Phenotype traits are then extracted using convex hull, meshing and computer-aided design (CAD) models. A HTPP for large scale 3D phenotyping of vineyards under field conditions is developed in \cite{ROSE}. It uses a track-driven vehicle equipped with a five camera system and a RTK-GPS to reconstruct a textured point cloud of a grapevine row based on a MVS algorithm. A classification algorithm is then employed to segment the plant into grape bunches and canopy, and extract phenotypic data such as quantity of grape bunches, berries and berry diameter. \\One big limitation of vision systems is the sensitivity to changing lighting conditions, which are common in outdoor applications like agriculture. In such situations, LiDAR sensors have been proved to be much more reliable, although they usually need higher investment costs. A LiDAR sensor mounted on a ground tripod is proposed in \cite{KEI} for 3D volumetric modeling of a grapevine under laboratory conditions. A ground fixed LiDAR is used in \cite{RAUM} to model the tree skeleton based on a graph splitting technique.  In \cite{AUAT}, a LiDAR sensing device carried by a tractor is proposed to scan fruit tree orchards as the vehicle navigates in the environment and acquire data for canopy volume estimation through a set of computational geometry methods. A ground-based LiDAR scanner is used in \cite{ARN} for leaf area index estimation in vineyards. HTPPs using field phenotyping robots have been also developed, for applications in maize \citep{RUC}, small grain cereals \citep{BUS} and viticulture \citep{SCHW,LON}. \\As an alternative to vision and LiDAR systems, portable consumer-grade RGB-D sensors, like Microsoft Kinect, have been recently proposed, mostly in controlled indoor environments such as greenhouses. By including hardware accelerated depth computation over a USB connection, these systems features a low-cost solution to recover in real-time 3D textured models of plants \citep{YAN}. In \cite{PAU} the Kinect and the David laser scanning system are analyzed and compared under laboratory conditions to measure the volumetric shape of sugar beet taproots and leaves and the shape of wheat ears. Depth images from a Kinect are used in \cite{CHEN} to segment and extract features representative of leaves of rosebush, yucca and apple plants in an indoor environment. In \cite{CHAI}, a Kinect is adopted for 3D reconstruction and phenotyping of corn plants in a controlled greenhouse setting. Depth and RGB cameras are mounted onboard a mobile platform for fruit detection in apple orchards \citep{GON}, whereas structural parameters including size, height, and volume are obtained for sweet onions and cauliflowers, respectively in \cite{WAN} and \cite{AND}.\\ Preliminary tests to estimate the grape mass using a Kinect sensor are presented in \cite{MAR}. An application of the Kinect sensor for three-dimensional characterization of vine canopy is also presented in \cite{MAR1}, using the sensor statically positioned in the vineyard. Although recent developments of Kinect have led to higher robustness to artificial illumination and sunlight, some filters need to be applied to overcome the increased noise effects and to reach exploitable results \citep{ROS}, therefore the application of this sensor remains mostly limited to indoor contexts and for close range monitoring \citep{LAC, ZEN}. \\In 2015, Intel introduced a novel family of highly portable, consumer depth cameras based on infrared stereoscopy and an active texture projector, which is similar to the Kinect sensor in scope and cost, but with a different working principle. The main advantage of this sensor with respect to Kinect is that it can work in zero lux environment as well as in broad daylight, which makes it suitable for outdoor conditions. In addition, its output include RGB information, infrared images and 3D depth data, thus covering a wide range of information about the scene. Hence, this sensor potentially provides an affordable and efficient solution for agricultural applications. Previous work has shown that the RealSense can be successfully used for 3D visual mapping tasks \citep{GAL}. \\To the best of our knowledge, the study presented in this paper is the first application of this kind of RGB-D sensor in the context of in-field grapevine phenotyping. Mounted on an agricultural tractor, the sensor is able to provide ground-based 3D information about the canopy volume. At the same time, using RGB images only it is possible to detect and count grape clusters. The main contribution of the proposed system is that data acquisition and processing can be carried out during vehicle operations, in a non-invasive and completely automatic way, requiring at the same time low investment and maintenance costs.
\\Another contribution of the proposed approach regards the use of a deep-learning based method for grape bunch recognition. In the last few years, deep learning \citep{LEC} has become the most promising technique of the broad field of machine learning for applications in the agricultural scenario \citep{KAM}. The capacity of deep learning to classify data and predict behaviors with high flexibility allows its use in fruit recognition and counting in orchards from visual images \citep{SA,BAR, RAH}.
In addition, deep-learning-based approaches have demonstrated high capability in plant identification \citep{LEE,GRIN}, enabling their use for weed-detection from aerial images \citep{DOS}. At the same time, several studies have been proposed for image-based plant disease classification \citep{FER,TOO}. The improved capabilities of deep learning in comparison with standard machine learning derive from the higher complexity of the networks, since phenomena (or corresponding data) under analysis are characterized by hierarchical representations.
Such representations are computed through the application of consecutive layers of filters, whose aggregation constitutes the artificial net. These filters create higher abstractions as data flows "deeper" in the net. Preliminary training of the net can automatically determine the best filters, i.e. the most significant and robust features, to achieve classification. \\In this work, a comparative study of four pre-trained convolutional neural networks (CNNs) designed to achieve the task of grape bunch recognition is reported. Following the paradigm of transfer learning \cite{YOS}, the four pre-trained CNNs, namely the AlexNet \citep{KRI}, the VGG16 and the VGG19 \citep{SIM}, and the GoogLeNet \citep{SZE}, are tuned by a lower number of samples of the training set, thus enabling image-based grape cluster detection.

\section{Materials and Methods}\label{s3}
\begin{figure}[]
      \centering
      \includegraphics[height=6 cm]{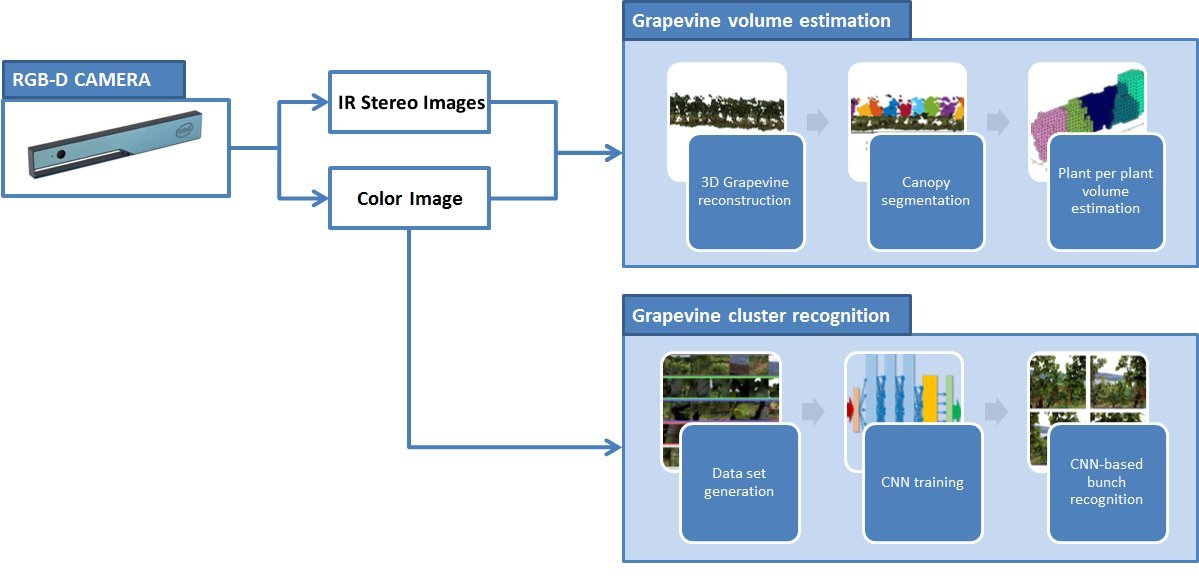}
      \caption{Pipeline of the proposed framework for in-field crop volume estimation and grape bunch detection.}
      \label{pipe}
\end{figure}

A sensing framework is proposed to automatically estimate the crop volume and detect grape bunches in vineyards, using a low-cost RGB-D sensor mounted on-board an agricultural vehicle. The overall processing pipeline is explained in Figure \ref{pipe}. The sensor provides an infrared (IR) stereo pair and a RGB image, which are fed to an algorithm for 3D reconstruction of a grapevine row and canopy segmentation from the trunks and other parts of the scene. The segmented canopy is successively processed using a set of computational geometry methods to estimate the volume of the plants. In addition, color images are used as input to a deep learning-based approach to recognize grape clusters.\\Data for algorithm development and testing were acquired in a commercial vineyard of a white variety of wine grape, named R{$\ddot{\textrm{a}}$}uschling, located in Switzerland.

\subsection{Acquisition platform}\label{s3.1}
\begin{figure}[]
      \centering
      \includegraphics[height=6 cm]{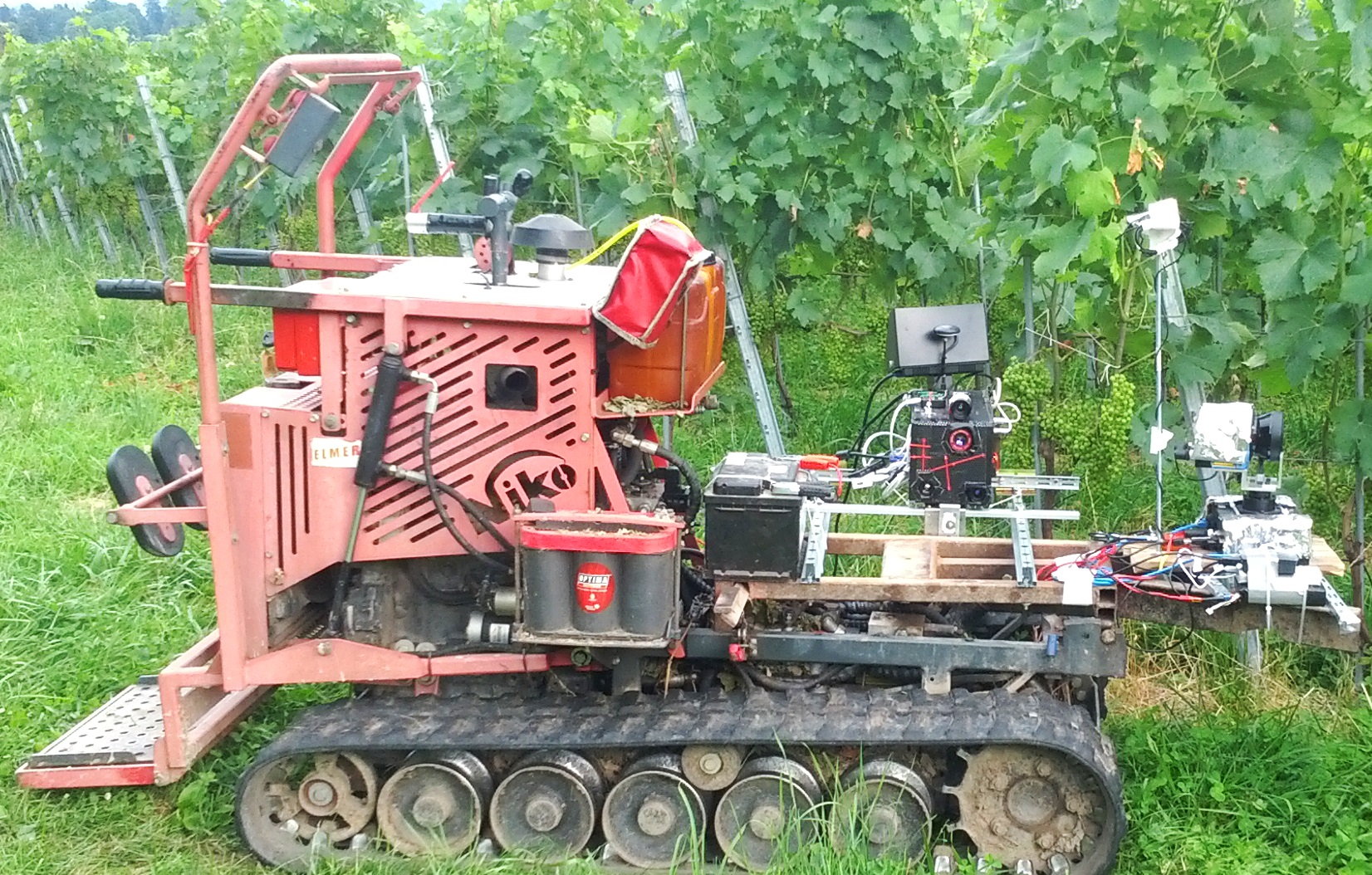}
      \caption{Niko caterpillar vehicle used for experimentation equipped with a multi-sensor suite.}
      \label{fig::vehicle}
\end{figure}

\begin{figure}[]
      \centering
     \subfigure[] {\includegraphics[height=5 cm]{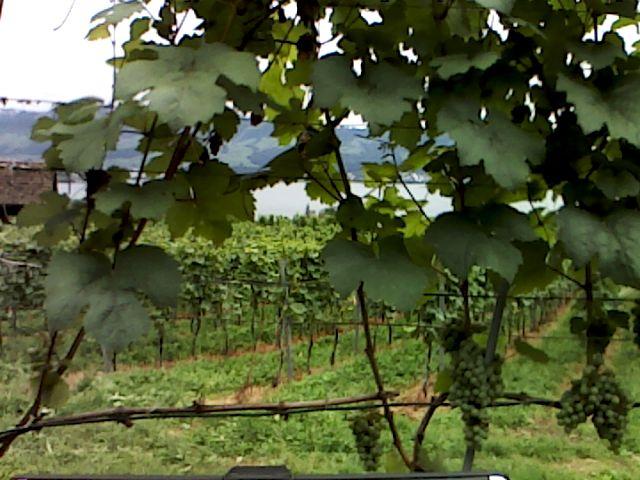}}
     \subfigure[] {\includegraphics[height=5 cm]{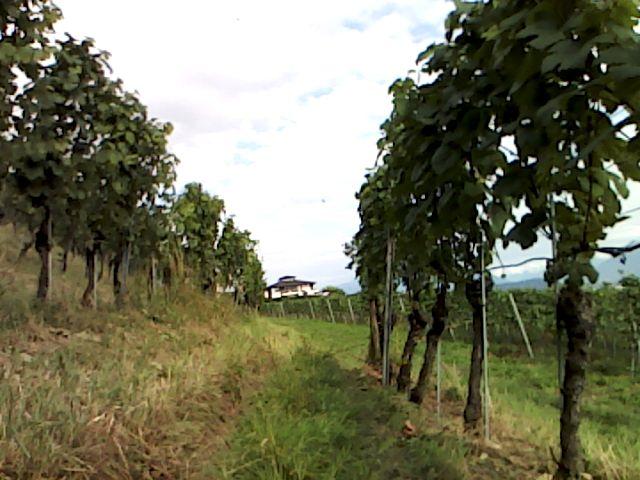}}
      \caption{Sample images acquired by the R200 camera, (a) mounted laterally and (b) mounted in a forward looking configuration.}
      \label{fig::sample_images}
\end{figure}

An Intel RealSense RGB-D R200 imaging system was used. It includes a left-right stereo pair and a color camera. The stereo pair consists of two global shutter VGA ($640\times480$) CMOS monochrome cameras with a visible cut filter at approximately 840 nm, with a nominal baseline of 70 mm and a nominal field of view of $60(H)\times45(V)\times70(D)$ deg. They run at 30, 60 or 90 Hz. The color camera is a FullHD ($1920\times1080$) Bayer-patterned, rolling shutter CMOS imager. Although, a global shutter approach may ensure better performance, it results in significantly higher cost and complexity. In addition, during the experimental campaign the CMOS imager proved to be effective with negligible rolling shutter type artifacts. The stream acquired by the color camera is time-synchronized with the stereo pair. Its nominal field of view is of $70(H)\times43(V)\times77(D)$ deg and runs at 30Hz at FullHD and higher frame rate for lower resolutions. The R200 sensor also includes an infrared texture laser-based (Class 1) projector with a fixed pattern, which helps improving image matching in texture-less surfaces, which is especially useful in indoor environments, where homogenous regions are more likely to occur.
\\The sensor was mounted on-board a commercial Niko caterpillar (see Figure \ref{fig::vehicle}). The camera was mounted first laterally, facing the side of the central-lower part of the canopy, where most of grape clusters can be typically found. A forward looking configuration was adopted instead for the task of volume estimation, due to the relatively limited vertical field of view of the camera and maximum allowed distance of the vehicle with respect to the row during the traversal, which did not make possible to frame the entire canopy of the plants when the camera was pointing laterally. Two sample images acquired using the lateral and forward looking camera configuration are shown in Figure \ref{fig::sample_images} (a) and (b) respectively.\\Images were acquired using the ROS \emph{librealsense} library. Due to limitations on the on-board storage capability of the acquisition systems, image resolution was limited to $640\times480$ for color images with a frame rate of 5 Hz. In addition, color images were compressed using the \textit{compressed\_image\_transport} plugin package that enables a JPG compression format.

\subsection{Canopy volume estimation}\label{s3.2}
The proposed approach for canopy volume estimation consists of three main phases:
\begin{enumerate}
    \item  3D grapevine row reconstruction;
    \item  grapevine segmentation;
    \item  plant per plant volume estimation.
\end{enumerate}

\subsubsection{3D grapevine row reconstruction}\label{s3.2.1}
\begin{figure}[]
      \centering
     \subfigure[] {\includegraphics[height=3.3 cm]{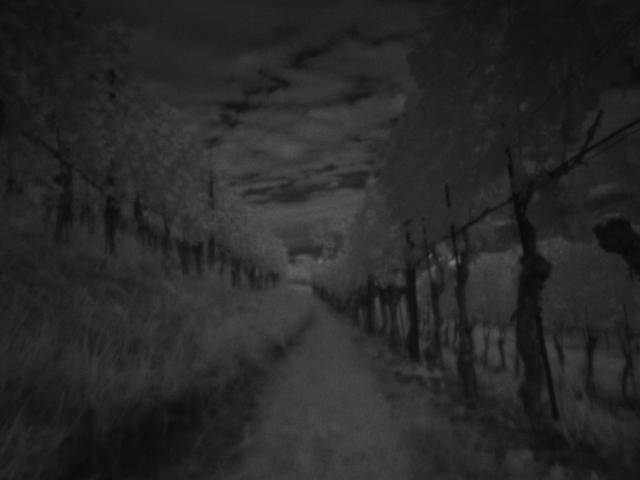}}
     \subfigure[] {\includegraphics[height=3.3 cm]{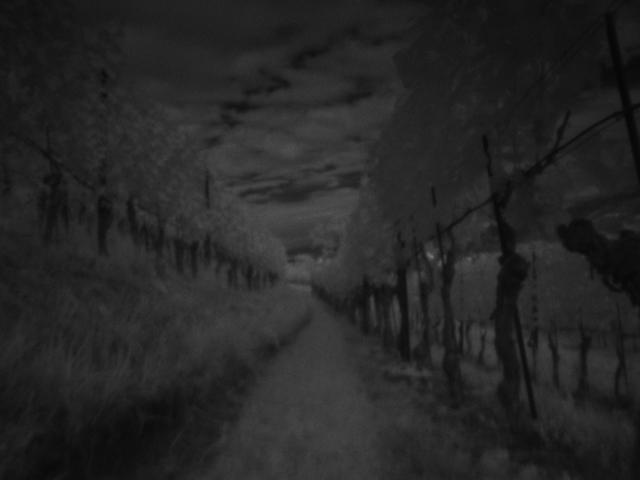}}
     \subfigure[] {\includegraphics[height=3.3 cm]{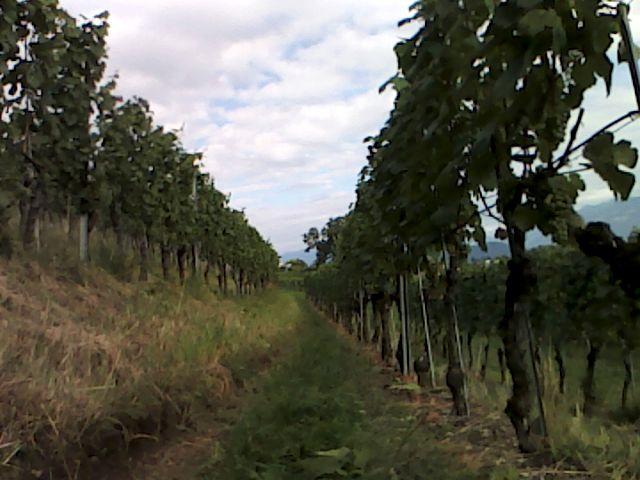}}\\
     \subfigure[] {\includegraphics[height=3.3 cm]{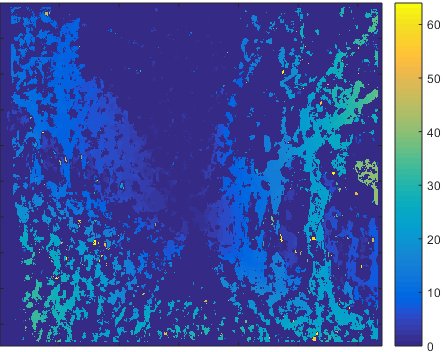}}
     \subfigure[] {\includegraphics[height=3.3 cm]{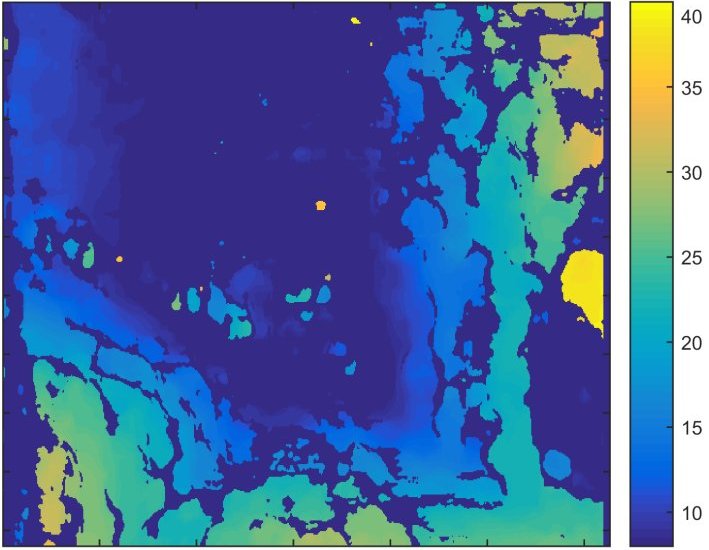}}
      \caption{(a)-(b) Left-right infrared images and (c) corresponding color image. (d) Disparity map obtained from the on-board sensor computation. (e) Result of the SGM algorithm with custom stereo parameters settings. }
      \label{fig::stereo_images}
\end{figure}

The first step of data processing consisted in building a color 3D point cloud representative of the whole grapevine row. The R200 features an on-board imaging processor, which uses a Census cost function to compare left and right images, and generate depth information. The algorithm is a fixed function block and most of its matching parameters are not configurable \citep{KES}. In particular, it uses a fixed disparity search range of 64 with a minimum disparity of 0, which, at the given resolution, corresponds to a range of reconstruction distances from about 0.65 m up to infinity. Instead, a disparity search range of [8, 40] pixel corresponding to a depth range of approximately 1.0 to 5.5 m was found to be a good choice for the task at hand. Therefore, the depth output by the camera was not employed in this study, and the infrared stereo pair was used as input to a stereo reconstruction algorithm running on an external computing device.
A calibration stage was preliminarily performed to align the IR-generated depth map with the color image, as well as to estimate the sensor position with respect to the vehicle, based on a calibration procedure using a set of images of a planar checkerboard\footnote{http://www.vision.caltech.edu/bouguetj/calib$\_$doc/}.
Then, for each stereo pair, the reconstruction algorithm proceeds as follows:
\begin{itemize}
      \item \emph{Disparity map computation}: to compute the disparity map, the Semi-Global Matching (SGM) algorithm is used. This algorithm combines concepts of local and global stereo methods for accurate, pixel-wise matching with low runtime \citep{HIR}. An example of raw images and corresponding disparity map is shown in Figure \ref{fig::stereo_images}. In particular, Figure \ref{fig::stereo_images}(a)-(b) show the infrared left-right images and Figure \ref{fig::stereo_images}(c) displays the corresponding color frame. Figure \ref{fig::stereo_images}(e) shows the disparity map computed by the SGM algorithm in comparison with the one streamed out by the sensor in Figure \ref{fig::stereo_images}(d). It can be seen that the latter provides a sparse reconstruction of the scene, including many regions not pertaining to the grapevine row of interest (i.e., the one on the right side of the picture), which could negatively affect the overall row reconstruction process. Instead, the disparity search range of the SGM algorithm was set to better match the region of interest in the scene. Additional stereo processing parameters, such as block size, contrast and uniqueness thresholds, were also configured as a trade-off of accuracy and density.
      \item \emph{3D point cloud generation}: being the stereo pair calibrated both intrinsically and extrinsically, disparity values can be converted in depth values and 3D coordinates can be computed in the reference camera frame for all matched points \citep{SER}. Only the points for which also color information is available from the color camera are retained, so as to obtain a 3D color point cloud. Point coordinates are successively transformed from camera frame to vehicle frame, using the transformation matrix resulting from the calibration process.
      \item \emph{Point selection}: a statistical filter is applied to reduce noise and remove outlying points. Furthermore, only the points located within a certain distance from the right side of the vehicle are selected. This allows to discard most of the points not pertaining to the grapevine row.
\end{itemize}

Subsequent point clouds are stitched together to produce a 3D map. To this end, the vehicle displacement between two consecutive frames is estimated using a visual odometry algorithm \citep{GEI}. A box grid filter approach is then applied for point cloud merging. Specifically, first, the bounding box of the overlapping region between two subsequent point clouds is computed. Then, the bounding box is divided into cells of specified size and points within each cell are merged by averaging their locations and colours. A grid step of 0.01 m was used as a compromise between accuracy and computational load. Points outside the overlapping region remain untouched.

\subsubsection{Grapevine segmentation}\label{s3.2.2}
In this phase, first, a segmentation algorithm is applied to separate the canopy from the trunks and other parts of the scene not pertaining to the canopy, such as the ground. Then, each single plant is detected using a clustering approach.
In order to detect the grapevine canopy, first of all, green vegetation is identified in the reconstructed scene based on the so-called Green-Red Vegetation Index (GRVI) \citep{TUC}. The GRVI is computed as:
\begin{equation}\label{eq::GRVI}
GRVI = \frac{\rho_{green}-\rho_{red}}{\rho_{green}+\rho_{red}}
\end{equation}
where $\rho_{green/red}$ is the reflectance of visible green/red. In terms of balance between green and red reflectance, three main spectral reflectance patterns of ground cover can be identified:
\begin{itemize}
\item green vegetation: where $\rho_{green}$ is higher than $\rho_{red}$;
\item soil (e.g., sand, silt, and dry clay): with $\rho_{green}$ lower than $\rho_{red}$;
\item water/snow: where $\rho_{green}$ and $\rho_{red}$ are almost the same.
\end{itemize}
According to Eq. \ref{eq::GRVI}, green vegetation, soil, and water/snow have positive, negative, and near-zero values of GRVI, respectively. It has been shown that $GRVI = 0$ can provide a good threshold value to separate green vegetation from other elements of the scene \citep{MOT}.
\\An algorithm for segmenting canopy from other parts of the scene was developed. First, the point cloud was divided into 3D grid cells of specified side. Then, for each cell the percentage $p$ of points with positive GRVI with respect to the total number of points, and the average height $\overline{H}$ above the ground were computed. Only the cells with $p>Th_p$ and $\overline{H}<Th_H$ were labeled as belonging to the grapevine canopy. For the given dataset, a value of $Th_H$ of 75 cm and $Th_p$ of 70\% were found to give good segmentation results.
\\Once the points pertaining to the row canopy have been selected, the single plants must be identified for the successive plant per plant volume estimation step. To this aim, the point cloud resulting from the canopy segmentation process was used to feed a \emph{K-means} algorithm.

\subsubsection{Plant per plant volume estimation}\label{s3.2.3}
Based on the results of the clustering algorithm, different computational geometry approaches were implemented to model the single plants and perform plant per plant volume estimation, as follows:
\begin{itemize}
  \item \emph{3D occupancy grid (OG)}: a box grid filter is applied, which divides the point cloud space into cubes or voxels of size $\delta$. The voxel size is a design parameter that is typically chosen as a trade-off between accuracy and performance \citep{AUAT}. Points in each cube are combined into a single output point by averaging their X,Y,Z coordinates, and their RGB color components. The volume of a plant is estimated as the sum of the volumes of all voxels belonging to the given plant foliage. This method is the one that ensures the best fitting of the reconstructed point cloud. It could lead however to an underestimation of the real volume in case of few reconstructed points, e.g., due to failure of the stereo reconstruction algorithm.
  \item \emph{Convex Hull (CH)}: this method consists in finding the volume of the smallest convex set that contains the points from the input cloud of a given plant \citep{DEB}. This method does not allow to properly model the concavity of the object, and could therefore lead to an overestimation of the volume.
  \item \emph{Oriented bounding box (OBB)}: this approach consists in finding the volume of the minimal bounding box enclosing the point set, \citep{ORU}. This method could lead to an overestimation of the plant volume as it is affected by the presence of outliers or long branches.
  \item \emph{Axis-aligned minimum bounding box (AABB)}: in this case the volume is defined as the volume of the minimum bounding box \citep{ORU} that the points fit into, subject to the constraint that the edges of the box are parallel to the (Cartesian) coordinate axes. As the OBB and CH approach, the AABB method can lead to an overestimation of the volume as it does not model properly plant voids and is affected by outliers or long branches.
\end{itemize}
The results obtained by each of these automatic methods were compared to the measurements obtained manually in the field at the time of the experimental test. In the manual procedure, the plant width is assumed to be constant and equal to the average distance between two consecutive plants (i.e., 0.9 m), whereas plant depth and height are measured with a tape; then, the volume is recovered as the product of width, depth, and height.

\subsection{Grape cluster recognition}\label{s3.3}
Grape cluster recognition is performed by exploiting four convolutional neural networks (CNN), namely the pre-trained AlexNet \citep{KRI}, VGG16 and VGG19 \citep{SIM}, and GoogLeNet \cite{SZE}. All networks allow for labeling input RGB images among 1000 different classes of objects, but with different architectures:
\begin{itemize}
  \item AlexNet: A feed-forward sequence of 5 convolutional layers and 3 fully-connected layers. The first convolutional layer has a size of 11x11, whereas the remaining have filter size of $3\times3$;
  \item VGG16: A feed-forward sequence of 13 convolutional layers and 3 fully-connected layers. All convolutional layers have filter sizes of $3\times3$;
  \item VGG19: A feed-forward sequence of 16 convolutional layers and 3 fully-connected layers. All convolutional layers have filter sizes of $3\times3$;
  \item GoogLeNet: A feed-forward sequence of 3 convolutional layers of $7\times7$, $1\times1$ and $3\times3$ filter sizes, 9 inception layers, each one made of 4 parallel blocks (without convolution and with $1\times1$, $3\times3$, $5\times5$ convolutional layers), and a single fully-connected layer at the end of the chain.
\end{itemize}
In all cases, the weights of the networks are determined during a training phase performed on a dataset of 1.2 million images.\\ Although these networks are devoted to solve the task of image labelling among generic classes, they can be tuned by keeping the first layers of the net and by rearranging the last fully-connected layer in order to solve more specific classifications. Under this condition, the whole architecture is almost completely initialized and, consequently, its parameters can be easily tuned by a smaller, but much more focused and target-oriented dataset.
In particular, here, the task of recognition is performed to classify input image patches, having size of the input layer of the CNNs ($227\times227$ for the AlexNet and $224\times224$  for the others), among 5 different classes of interest:
\begin{itemize}
  \item grapevine bunches;
  \item vineyard grape poles;
  \item grapevine trunks, cordons and canes (wood);
  \item leaves;
  \item background.
\end{itemize}

\begin{figure}[]
      \centering
      \includegraphics[width=13 cm]{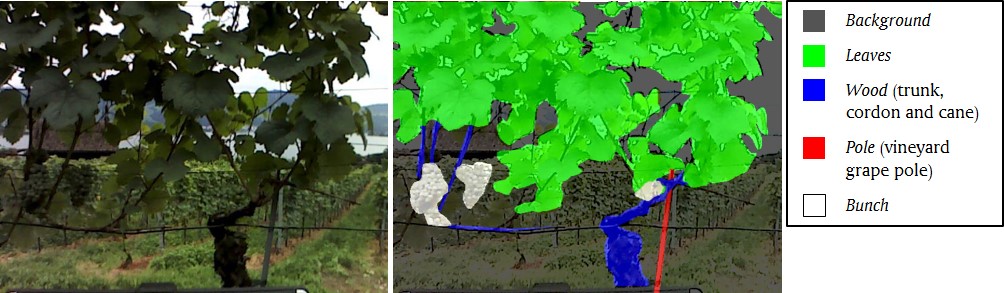}
      \caption {Example of an input image and the corresponding manual segmentation in five classes of interest: white, red, blue, green and black refer to Bunch, Pole, Wood, Leaves and Background, respectively.}
      \label{fig::segmentation_example}
\end{figure}

An example of segments extracted from a labelled input image is reported in Figure \ref{fig::segmentation_example}. It is worth noticing that the background class includes regions of pixels which are actually very similar to those of the other classes. For instance, leaves of far plants are almost similar to grape bunches of the foreground. The network shall discriminate between similarities to reduce false positives or false negatives.
\\Given the number of output classes, the starting CNNs have been changed by substituting the last fully-connected layer with another one made of only 5 neurons, one for each expected output class. This layer is then followed by a \emph{softmax} layer to compute classification scores. This fully-connected layer is not initialized and its weights have to be selected from scratch. Therefore, training the transfer nets requires different learning rates to keep the features from the early layers of the pre-trained networks. Specifically, the learning rate of the fully-connected layer is 20 times higher than the one of the transferred layer weights: this choice speeds up learning in the new final layer.
Training options are finally selected to run the learning phase on a single GPU (nVidia Quadro K5200). For instance, input data is divided in mini-batches, whose size is defined accordingly with the number of unknowns of the CNNs and the capability of the GPU (see Table \ref{tab::epochs_time} for details). At the end of each iteration the error gradient is estimated and the network parameters are updated in accordance with the principles of stochastic gradient descent with momentum (SGDM) optimization \citep{SUT}. Training is performed using two different datasets: the training and the validation sets. The former is used to estimate the network parameters by backpropagating the errors; the latter is proposed to the net at the last iteration of each epoch, in order to compute the validation accuracy and validation loss in feed forward for the current network parameters. If the validation loss does not go down for 10 consecutive epochs, the training of the network ends.

\subsubsection{Dataset generation}\label{s3.3.1}
\begin{figure}[]
      \centering
      \includegraphics[width=10 cm]{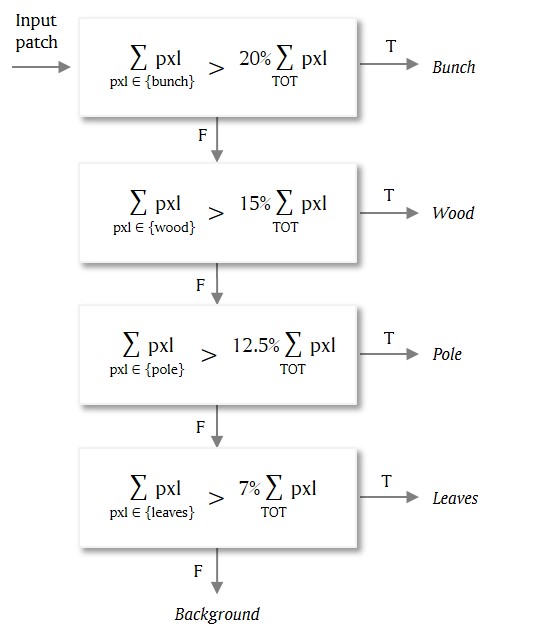}
      \caption{Flow chart for labeling of input training patches among the five output classes.}
      \label{fig::label_flow}
\end{figure}

\begin{figure}[]
      \centering
      \includegraphics[width=10 cm]{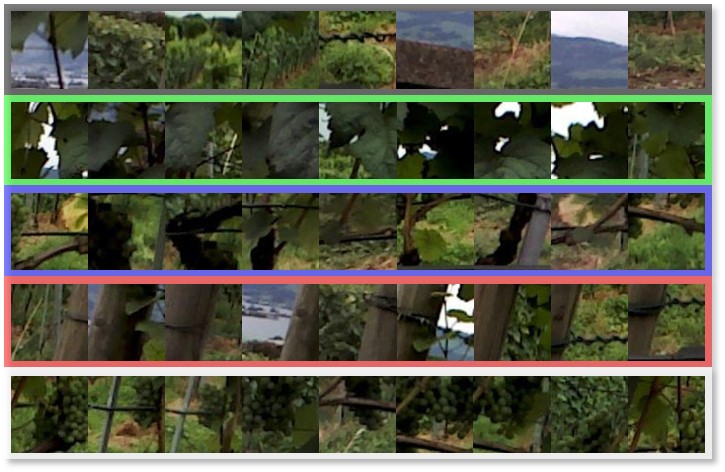}
      \caption{Sample patches used to fine-tune the CNNs. Black, green, blue, red and white frames refer to background, leaves, wood, pole and bunch classes, respectively.}
      \label{fig::patches_example}
\end{figure}

As previously reported, each CNN is designed to receive input images of N $\times$ N size (N = 227 for AlexNet and N = 224 for the other CNNs) and to output labels in a set of 5 possible classes. However, color frames of 640$\times$480 pixels were available from the R200 sensor. The whole set of visual images acquired during experiments and manually segmented in the 5 classes of interest is made of 85 images. This set is randomly divided in two datasets with a proportion of 0.7-0.3. The first set is used for training purposes and is further separated in the training and validation sets. The second set is made of test images, which will be proposed to the network after its training, in order to compute quality metrics. These images are not known to the network which will proceed to labelling.\\
The target of image analysis is the detection and localization (segmentation) of grape bunches within the input images. This problem can be solved by moving a square window, whose size is large enough to frame bunches, and by processing each resulting patch. Then, depending on the classification score, grape clusters can be detected and thus localized within the images. Specifically, grape bunches have sizes which depend on the camera field-of-view and the plant distance. Under the current working conditions, bunches can be framed in windows whose sizes do not exceed 80$\times$80 pixels. Accordingly, the proposed algorithm processes input images by moving a 80-by-80-window at discrete steps, in order to extract image patches. This process of patch extraction is speeded up by means of a pre-determined look-up table, filled by the indices, i.e. memory positions, of each entry of the expected patches. These windowed images will then feed the network in both training and test phases. Before feeding the network, patches are upscaled using bicubic interpolation to the target size of N$\times$N, in order to match the input layer of the corresponding CNNs. It is worth noticing that bicubic interpolation can add artifacts to input images, such as edge rounding, which can down the accuracy of classification results \citep{DOD}. However, all images of both training and test datasets are upscaled to the same patch size and, thus, upscaling distortions equivalently affect both datasets. As a consequence, the networks are trained by distorted images and tune their weights to process test images with the same distortions of the training one. As a result, classification will not be altered by upscaling distortions, since the networks are already trained to manage them.
Examples for training the network are created by labelling each image patch of the training and validation sets depending on the number of pixel belonging to a specific class. In this case, the choice is made by a weighted summation pixel labels, in order to give more significance to pixels of bunch, pole and wood classes. The simple flow chart for patch labeling is reported in Figure \ref{fig::label_flow}.
For instance, if the summation of pixels of the bunch class goes beyond 20\% of the whole number of pixels of the input patch (more than 1280 over 6400 pixels, before upscaling), the example trains the network as bunch. Moving the window and counting the pixels of the ground truth generates 44400, 5295, 3260, 7915 and 81687 patches of the classes leaves, bunch, wood, pole and background, respectively. A subset of patches of the training and validation sets is presented in Figure \ref{fig::patches_example}.

\section{Results and Discussion}\label{s4}
\begin{figure}[]
      \centering
      \includegraphics[height=6 cm]{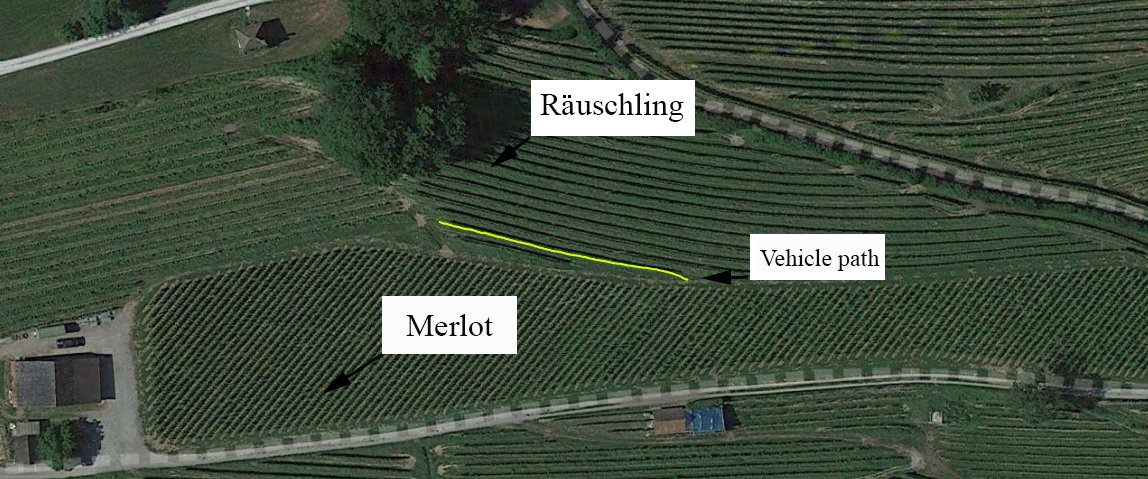}
      \caption{Google Earth view of the testing environment in a commercial vineyard in Switzerland. It comprises two types of grapes, namely R{$\ddot{\textrm{a}}$}uschling and Merlot. The yellow line denotes the vehicle path as estimated by an on-board RTK GPS system during the experiment.}
      \label{fig::map}
\end{figure}

Field tests were performed in a commercial vineyard in Switzerland. A picture showing a Google Earth view of the testing environment is reported in Figure \ref{fig::map}. The vehicle trajectory estimated by an on-board RTK GPS system is overlaid on the map in Figure \ref{fig::map}. Data were acquired by driving the vehicle from left to right at an average speed of 1.5 m/s. 640$\times$480 stereo frames and corresponding color images were captured at a frame rate of 5 Hz. The row to the right side in direction of travel, including 55 plants, is considered in this study. Note that the fifth plant in the direction of motion was judged to be dead by visual inspection and was not considered in this work, resulting in a total of 54 surveyed plants.
\subsection{Canopy volume estimation}\label{s4.1}
\begin{figure}[]
      \centering
      \includegraphics[width=\columnwidth]{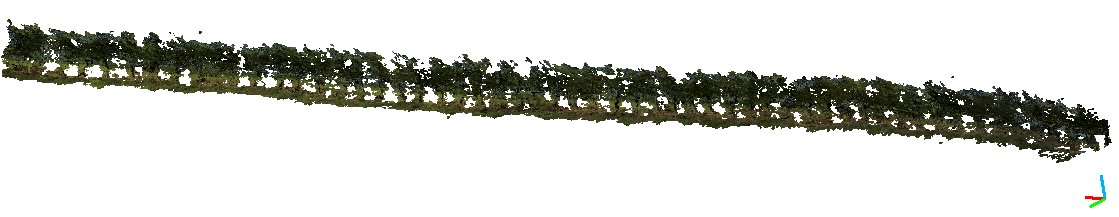}
      \caption{3D reconstruction of the grapevine row under inspection. It includes 55 plants spanning along a total length of approximately 50 m. Plants are located at approximately 0.90 m apart with respect to each other.}
      \label{fig::3D_rec}
\end{figure}

\begin{figure}[]
      \centering
      \includegraphics[width=\columnwidth]{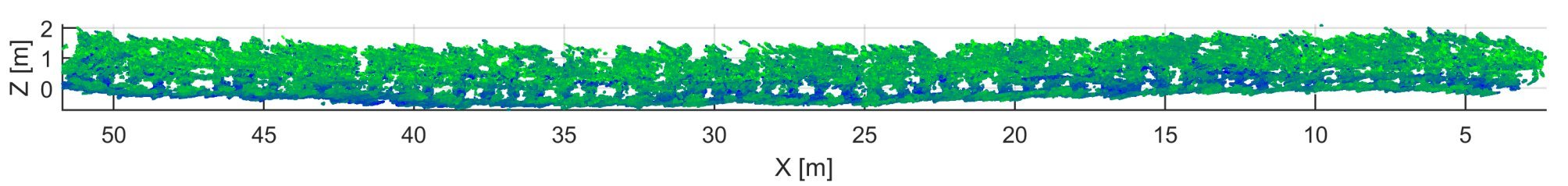}
      \caption{GRVI map: green points correspond to green vegetation; blue points correspond to non-vegetated regions. Lighter green corresponds to higher GRVI.}
      \label{fig::GRVI_map}
\end{figure}

\begin{figure}[]
      \centering
      \includegraphics[width=\columnwidth]{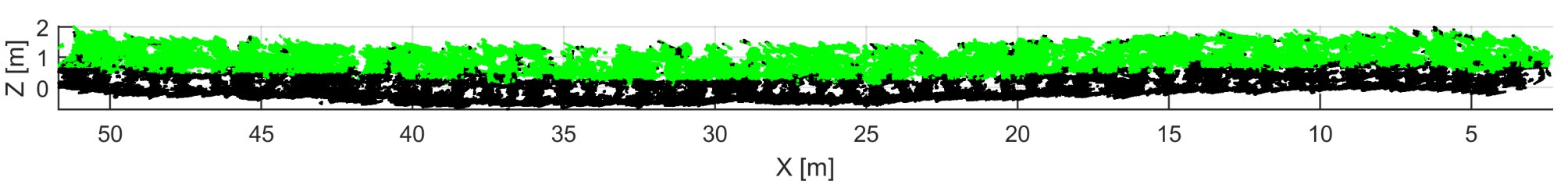}
      \caption{Segmentation results: green points pertain to foliage, whereas black points belong to other parts of the scene.}
      \label{fig::canopy}
\end{figure}

The output of the 3D reconstruction method for the grapevine row under analysis is reported in Figure \ref{fig::3D_rec}. It comprises about 1.3M points. The corresponding GRVI map is shown in Figure \ref{fig::GRVI_map} where lighter green denotes higher values of GRVI. The result of the segmentation algorithm using GRVI and 3D information is shown in Figure \ref{fig::canopy}: points displayed in green pertain to the plant foliage, whereas black points are classified as non-leaf points and are not considered for the subsequent volume estimation procedure.

\begin{figure}[]
      \centering
      \includegraphics[width=\columnwidth]{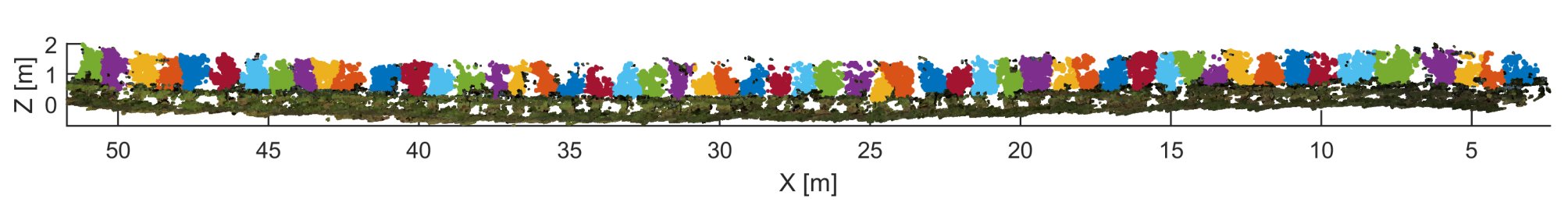}
      \caption{Result of the clustering process. Single plants are represented by a different color, for a total of 54 plants. Non-leaf points are displayed in their original RGB color.}
      \label{fig::clusters}
\end{figure}

\begin{figure}[]
      \centering
     \subfigure[] {\includegraphics[height=4 cm]{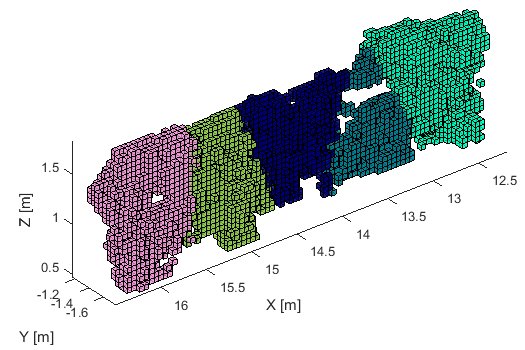}}
     \subfigure[] {\includegraphics[height=4 cm]{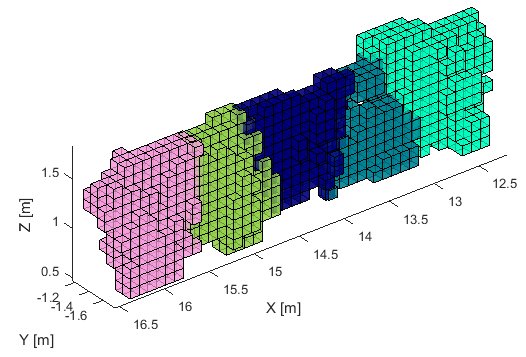}}
     \subfigure[] {\includegraphics[height=4 cm]{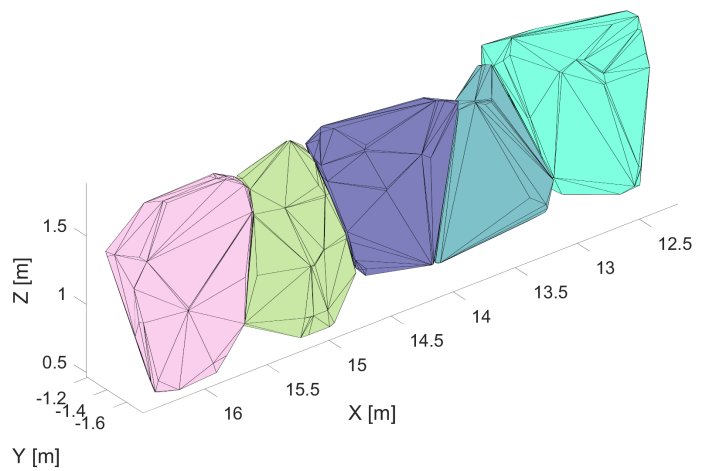}}
     \subfigure[] {\includegraphics[height=4 cm]{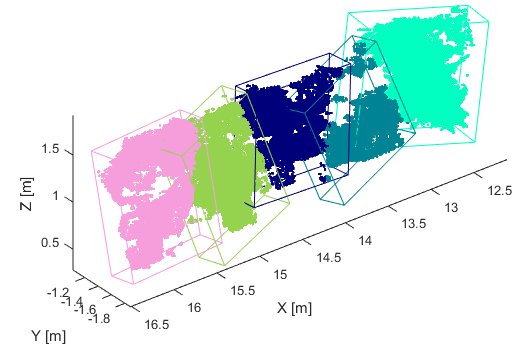}}
     \subfigure[] {\includegraphics[height=4 cm]{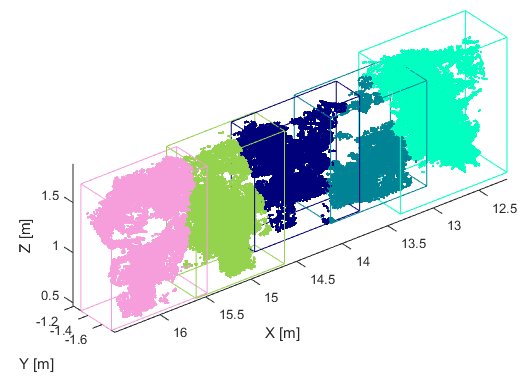}}
      \caption{Result of plant modeling for plants from 10th to 14th using 3D occupancy grid with $\delta$ = 0.05 m (a) and $\delta$ = 0.1 m (b),  convex hull (c), oriented bounding box (d) and axis-aligned minimum bounding box (e). }
      \label{fig::sample_vols}
\end{figure}

\begin{figure}[]
      \centering
      \includegraphics[width=12 cm]{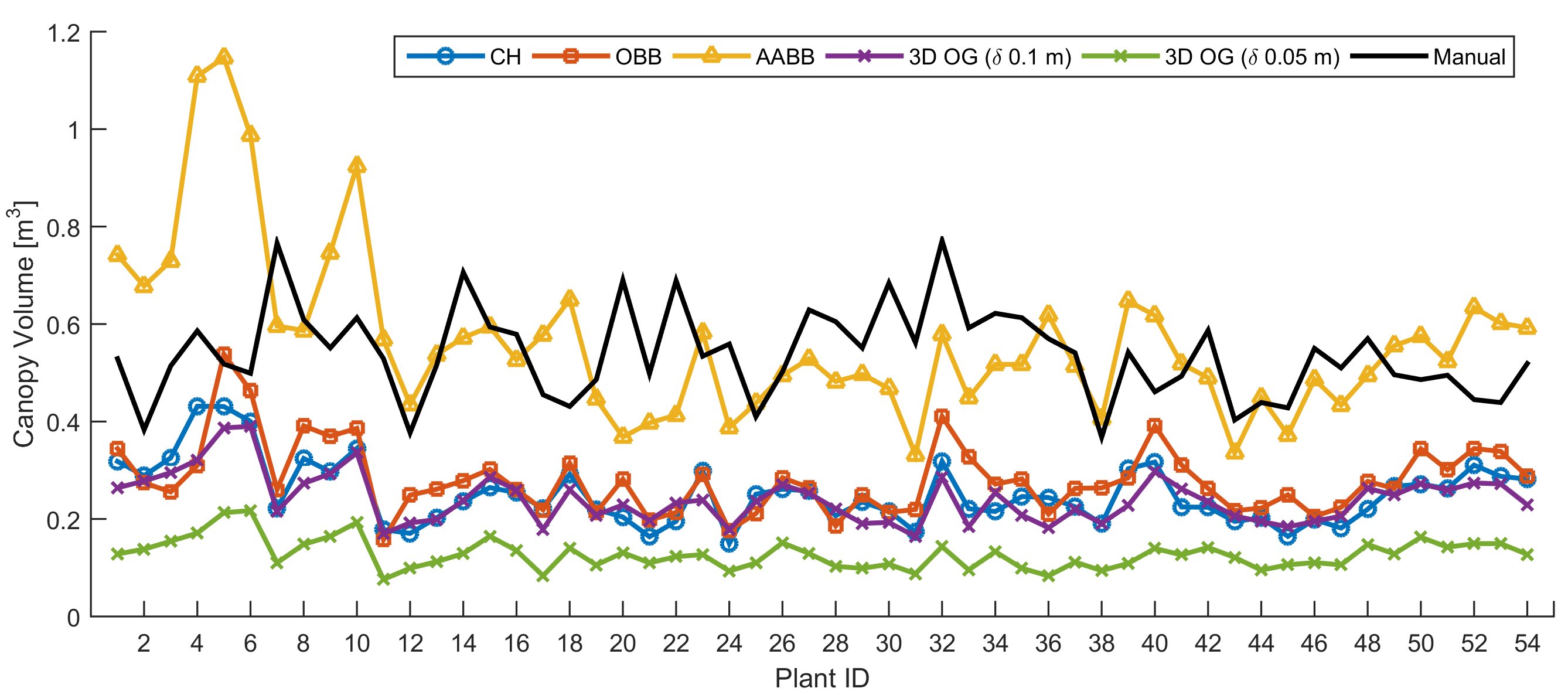}
      \caption{Plant volume estimates obtained using different automatic methods compared with a manual approach.}
      \label{fig::volumes}
\end{figure}

\begin{figure}[]
      \centering
      \includegraphics[width=14 cm]{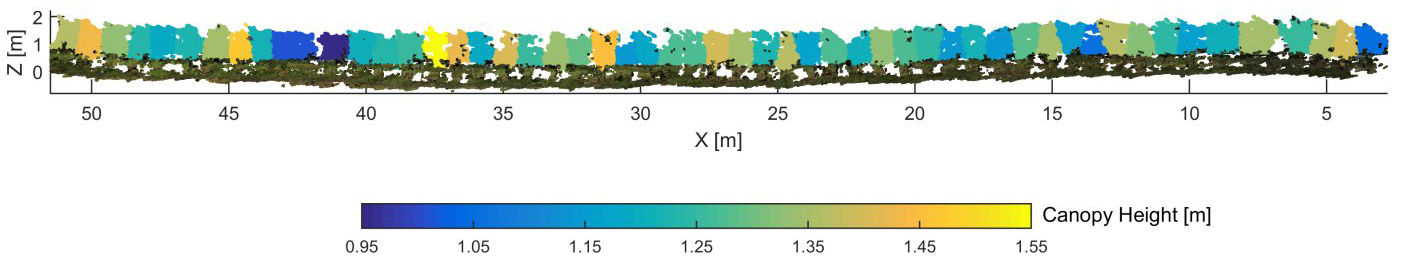}
      \caption{Canopy height estimated as the vertical extension of the minimum bounding box of individual plants. The color of each plant reflects the canopy height according to the color bar shown at the bottom of the figure.}
      \label{height}
\end{figure}

Points classified as pertaining to plant foliage were further processed by \emph{k-means} clustering. The number of clusters was fixed at 54. Centroids were initialized knowing that plants are approximately located at an average distance of 0.90 m with respect to each other. Figure \ref{fig::clusters} shows the result of the clustering process. Each cluster is represented by a different color. Non-leaf points are also displayed in their original RGB color. For each cluster, the volume was estimated using the four methods described in Section \ref{s3.2.3}, and was compared with the manual measurements. \\Figure \ref{fig::sample_vols} shows, as an example, the result of plant modeling using the different methods for five individuals. Estimated volumes for all individuals are shown in Figure \ref{fig::volumes}. The corresponding descriptive statistics in terms of mean, standard deviation, minimum and maximum measure are reported in Table \ref{tab::stats}. These results show that methods not taking into account concavities and voids such as CH, OBB, and AABB, result in higher estimates, with the AABB method most approaching the manual measurements with an average discrepancy of about 5\%. A significantly smaller volume is obtained using the 3D-OG approach with a voxel grid size $\delta = 0.05$ m, due to disconnected structures and voids in the model. This effect is mitigated when increasing the grid size up to $0.1$ m, with results getting closer to the CH and OBB methods.
\begin{table}[]
  \centering
  \caption{Descriptive statistics of canopy volume ($m^3$) by different methods.}
  \label{tab::stats}
  \begin{tabular}{lccccc}
     & Mean  & St. Dev. & Min.& Max.\\
    METHOD &($m^3$) & ($m^3$) &($m^3$) & ($m^3$)\\
    \hline
    OG ($\delta=0.05 m$)& 0.127&	0.031&	0.076&	0.217\\
    OG ($\delta=0.1 m$) &0.241	&0.050	&0.164	&0.390\\
    CH & 0.252 &	0.064 &	0.150	&0.431\\
    OBB & 0.281&	0.072&	0.158&	0.538\\
    AABB & 0.564	&0.169	&0.331	&1.147\\
    Manual& 0.539	&0.091	&0.368	&0.771\\
  \end{tabular}
\end{table}\\
As a final remark, it should be noted that the availability of the 3D point cloud pertaining to each single plant makes it possible to extract other geometric traits of interest. As an example, Figure \ref{height} shows the estimation of the canopy height, obtained as the vertical extension of the minimum bounding box for the same vineyard row. Individual plants are coloured based on the estimated canopy height according to the color bar shown at the bottom of Figure \ref{height}. Non-leaf points are also displayed in their original RGB color. The average canopy height resulted in 1.25 m with a standard deviation of 0.12 m.

\subsection{Grape cluster detection}\label{s4.2}
The pre-trained CNNs have been fine-tuned to perform the classification among the five classes of interest, through proper training. Then, they have been used to label test patches and to compare final scores in prediction tasks. This section describes the evaluation metrics for the assessment of results, the training phase on labelled images and the final tests on unknown data.

\subsubsection{Evaluation metrics}\label{s4.2.1}
As already shown in the previous sections, the CNNs receive square image patches of N$\times$N pixels (N = 227 for AlexNet and N = 224 for VGG16, VGG19 and GoogLeNet). Each input patch of both training and test sets has a corresponding label, among the five target classes, determined by manual segmentation of the dataset. Accordingly, it is possible to define some metrics which will be helpful for the final evaluation of the net performance. With reference to multiclass recognition, for each class $c$, the final results of training will be presented by counting true and false predictions of class $c$ against the others. It leads to true positives ($TP_c$) and negatives ($TN_c$), and false positives ($FP_c$) and negatives ($FN_c$). These values are further processed to determine accuracy ($ACC_c$), balanced accuracy ($BACC_c$), precision ($P_c$), recall ($R_c$), and true negative rate ($TNR_c$), which have the following formulations:

\begin{equation}
ACC_c = \frac{TP_c+TN_c}{TP_c+FP_c+TN_c+FN_c}\label{eq1}
\end{equation}
\begin{equation}
BACC_c = \frac{\frac{TP_c}{TP_c+FN_c}+\frac{TN_c}{TN_c+FP_c}}{2}\label{eq2}
\end{equation}
\begin{equation}
P_c = \frac{TP_c}{TP_c+FP_c}\label{eq3}
\end{equation}
\begin{equation}
R_c = \frac{TP_c}{TP_c+FN_c}\label{eq4}
\end{equation}
\begin{equation}
TNR_c = \frac{TN_c}{TN_c+FP_c}\label{eq5}
\end{equation}
Specifically, $ACC_c$ estimates the ratio of correct predictions over the whole occurrences, while $BACC_c$ also weights standard accuracy by the number of occurrences of the $c^{th}$ class. As a consequence, $BACC_c$ overcomes misleading interpretations of results due to the heavy unbalancing of occurrences of the $c^{th}$ class in comparison to the other classes. $P_c$ shows whether all detections of the $c^{th}$ class are true, whereas $R_c$ states the ability of the CNN to label a patch of the $c^{th}$ class, recognizing it against the others. Finally, $TNR_c$ inverts the meaning of $R_c$, i.e. it defines how the CNN is capable of not assigning the label $c$ to patches of the other classes.
Metrics in Eq. \ref{eq1} to \ref{eq5} refer to pixel-by-pixel, i.e. per-patch, evaluations. Specifically, they can be effectively used for stating whether the architecture of the CNNs and the input examples are suitable to allow the classification of the training patches (training assessment). On the other hand, the result of tests will be evaluated to measure the actual capability of the CNN to detect grape clusters in 25 test images, knowing the number of bunches in each image, instead of evaluating the results of per-pixel segmentation. In this case, accuracy ($ACC_{GC}$), precision ($P_{GC}$) and recall ($R_{GC}$) are redefined as:
\begin{equation}\label{eq7}
ACC_{GC} = \frac{T_{GC}}{GC}
\end{equation}
\begin{equation}\label{eq8}
P_{GC} = \frac{T_{GC}}{T_{GC}+F_{GC}}
\end{equation}
\begin{equation}\label{eq9}
R_{GC} = \frac{T_{GC}}{T_{GC}+N_{GC}}
\end{equation}
where $GC$ is the total number of grape clusters, $T_{GC}$ is the number of true detections, $F_{GC}$ is the number of wrong detections and $N_{GC}$ is the number of missed detections of grape clusters.

\subsubsection{Training of the CNNs}\label{s4.2.2}
The training phase is performed by feeding the CNNs with 142000 sample patches, which are divided into training and validation sets in proportion 0.75-0.25. As previously stated, two different learning rates have been selected for the first layers, transferred from the CNNs, and for the last fully-connected layers. Specifically, the learning rate for the first layers is set to $10^{-4}$, whereas the last fully-connected layer, responsible for the final classification in the five classes of interest, is trained with a learning rate of $2\cdot10^{-3}$. The SGDM optimization is reached with a momentum of 0.9. These parameters produce the results in Table \ref{tab::epochs_time}, which reports the required number of epochs for reaching the convergence of the learning phase, and the corresponding time consumption (total and per-epoch), valid for the corresponding mini-batch size, whose dimension is defined to work at the memory limit of the GPU used for experiments.
\begin{table}[]
  \centering
  \caption{Number of epochs and time requirements for the learning phase of the four proposed CNNs.}
  \label{tab::epochs_time}
  \begin{tabular}{lccccc}
     & Mini-batch & Number of & Total training & Training time\\
     CNN & size & epochs & time &  per epoch\\
      & (\emph{no. of samples}) &  & (\emph{h}) & (\emph{min})\\
     \hline
    AlexNet & 210 &	17	& 4.0 &	14.1\\
    VGG16 & 35	& 15 &	57.5 &	230\\
    VGG19 & 10	& 11	& 49.1	& 268\\
    GoogLeNet & 21	& 22	& 20.2	& 55.1
  \end{tabular}
\end{table}

The analysis of Table \ref{tab::epochs_time} shows that the AlexNet, which is significantly less complex than any other CNN of this comparison, can be trained in 4 hours. On the contrary, the other CNNs require more than about 20 hours (see the GoogLeNet) to complete the learning phase.
Quantitative results on the network performance are in Table \ref{tab::table2}, which shows the evaluation metrics in Eq. \ref{eq1} to \ref{eq5}, computed by joining both the training and validation sets.

\begin{table}[]
  \centering
  \caption{Results of training. These metrics are computed based on Eq. \ref{eq1} to \ref{eq5} for the whole training set, made of both training and validation patches.}
  \label{tab::table2}
  \begin{tabular}{llcccccc}
    CNN & Class & $ACC{_c}$ & $BACC{_c}$& $P{_c}$& $R{_c}$ & $TNR{_c}$\\
     \hline
     & Background & 0.96054 &	0.96176	& 0.97714 &	0.95344	& 0.97007\\
     & Leaves & 0.97264 &	0.9726 &	0.94160 &	0.97248 &	0.97272\\
    AlexNet & Wood & 0.99508	& 0.94131 &	0.89819	& 0.88497 &	0.99765\\
     & Pole & 0.99527	& 0.98109	 & 0.95048	 & 0.96513	 & 0.99704\\
     & Bunch & 0.99347	& 0.97482 & 0.87974	 & 0.95467	& 0.99497\\
     \hline
      & Background & 0.99364 & 0.99362 & 0.99356 & 0.98586 &	0.99368\\
     & Leaves & 0.99888	& 0.99291	& 0.98646	& 0.98368	& 0.99937 \\
    VGG16  & Wood & 0.99882 &	0.98907 &	0.97882 &	0.97261 &	0.99932\\
     & Pole & 0.99866 &	0.99744	& 0.99607	& 0.98002	& 0.99881\\
     & Bunch & 0.99080	& 0.99113	& 0.98892	& 0.99507	& 0.99335\\
     \hline
     & Background & 0.9919	&0.9897	&0.98399	&0.98957	&0.9954\\
     & Leaves & 0.99623	&0.99642	&0.99661	&0.91078	&0.99622\\
    VGG19 & Wood & 0.99795	&0.98169	&0.96461	&0.95136	&0.99878 \\
     & Pole & 0.99863	&0.99426	&0.98934	&0.98596	&0.99918\\
     & Bunch & 0.98605	&0.98608	&0.98589	&0.98985	&0.98626\\
     \hline
     & Background & 0.98916	&0.99067	&0.99461	&0.9708	&0.98674\\
     & Leaves & 0.99732	&0.96642	&0.93303	&0.99479	&0.99981\\
    GoogLeNet & Wood & 0.9966	&0.99189	&0.98695	&0.88571	&0.99684\\
     & Pole & 0.99873	&0.99491	&0.99061	&0.98648	&0.99921\\
     & Bunch &0.98370	&0.98460	&0.97873	&0.99289	&0.99047
  \end{tabular}
\end{table}

The inspection of the results in Table \ref{tab::table2} reveals that the balanced accuracy is always higher than 94.13\% for all the CNNs (see the balanced accuracy of class wood of the AlexNet). With respect to the bunch class, the training phase of every proposed CNN ends with satisfying values of $BACC_{bunch}$, $P_{bunch}$ and $R_{bunch}$. Among these values, precision is comparable, but always lower than recall. It suggests that bunches will be detected with high rates, but among these detections several false positives could arise. For this reason, further processing, such as morphological filters on per-pixel segmentation, will be required to improve results in grape cluster detection.

\begin{figure}[]
      \centering
      \includegraphics[width=\columnwidth]{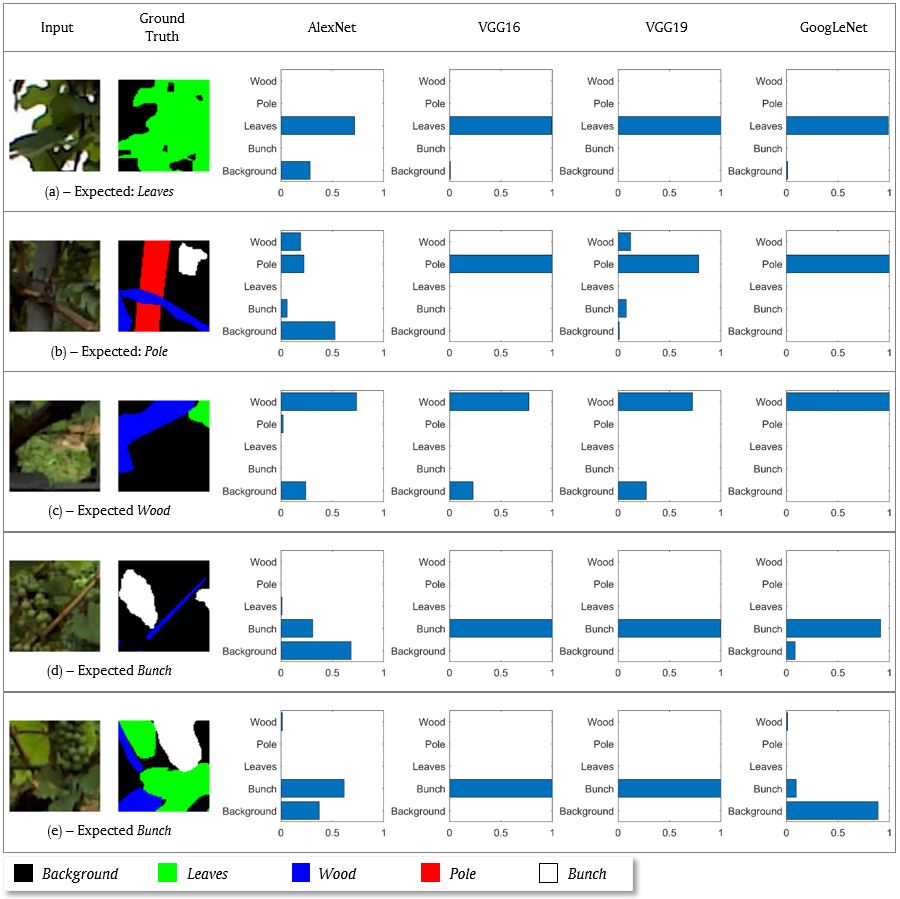}
      \caption{Results of tests on sample patches. Bar plots report the probabilities (classification scores) of image labels for the corresponding patches, by varying the CNN (AlexNet, VGG16, VGG19 and GoogLeNet). Pixel-by-pixel ground truths are reported for each pixel of the patches. Black, green, blue, red and white refer to regions of background, leaves, wood, pole and bunch, respectively. Patch labels are: (a) leaves, (b) pole, (c) wood and (d)-(e) bunch.}
      \label{fig::test_example}
\end{figure}

\subsubsection{Test of the CNN}\label{s4.2.3}
A test set of 25 images was used for evaluation purposes. Following the same approach developed for the training set, images are separated in smaller square patches, whose size is as large as the expected dimension of grape clusters (80$\times$80 pixels). Then, these patches are upscaled in order to meet the requirements of the input layer of the transferred CNNs. Each patch is finally labelled by the class of the highest relevance.
\\Test patches are used to feed the CNN, which returns 5 classification scores about the image label. As an example, Figure \ref{fig::test_example} shows the results of patch labelling in ambiguous cases, where the same patch has objects of different classes. As can be seen, each object (i.e., bunches, leaves, wood pieces and poles) contributes to increase the score of the corresponding class. In addition, as effect of the strategy adopted to label training patches, i.e. labelling the patches giving more weight to bunches, wood pieces and poles with respect to leaves and background, small parts of such objects have more influence to the final estimation of class probability. With reference to Figure \ref{fig::test_example}(d) and (e), few grapes surrounded by many leaves and/or large background areas will anyway produce the rise of the probability (score) of the bunch class.

This aspect is emphasized for the VGG16 and VGG19, which label the bunch patches with the highest probability (equal to 1), although patches in  \ref{fig::test_example}(d) and (e) have background, leaves and wood regions of extended area. On the contrary, the AlexNet and the GoogLeNet can fail in labeling bunches, as clearly shown by the inspection of corresponding probability values in Figure \ref{fig::test_example}(d) and (e), respectively. In both cases background scores are higher than those of the bunch class, expected for the two inputs.
Performance can be compared also in terms of time requirements for the complete classification of an input image. These requirements are reported in Table \ref{tab::time}, which shows the time elapsed for the classification of 12314 patches, taken from a single frame.  Time requirements follow the same behavior of the training phase. Specifically, the VGG19 takes more time with respect to the others, whereas the AlexNet, which has the simplest architecture, is the fastest CNN in labeling the whole image. However, this performance, which is obtained without implementing dedicated hardware, does not allow the online cluster recognition. As already discussed, each patch is taken from the input image at specific positions. Its processing produces 5 classification scores.

\begin{table}[]
  \centering
  \caption{Time elapsed for the classification of 12314 patches extracted from a single input image.}
  \label{tab::time}
  \begin{tabular}{cc}
      CNN & Time (s)\\
     \hline
    AlexNet & 27.18\\
    VGG16 & 417.56\\
    VGG19 & 521.84\\
    GoogLeNet & 331.40
  \end{tabular}
\end{table}

\begin{figure}[]
      \centering
      \includegraphics[width=12 cm]{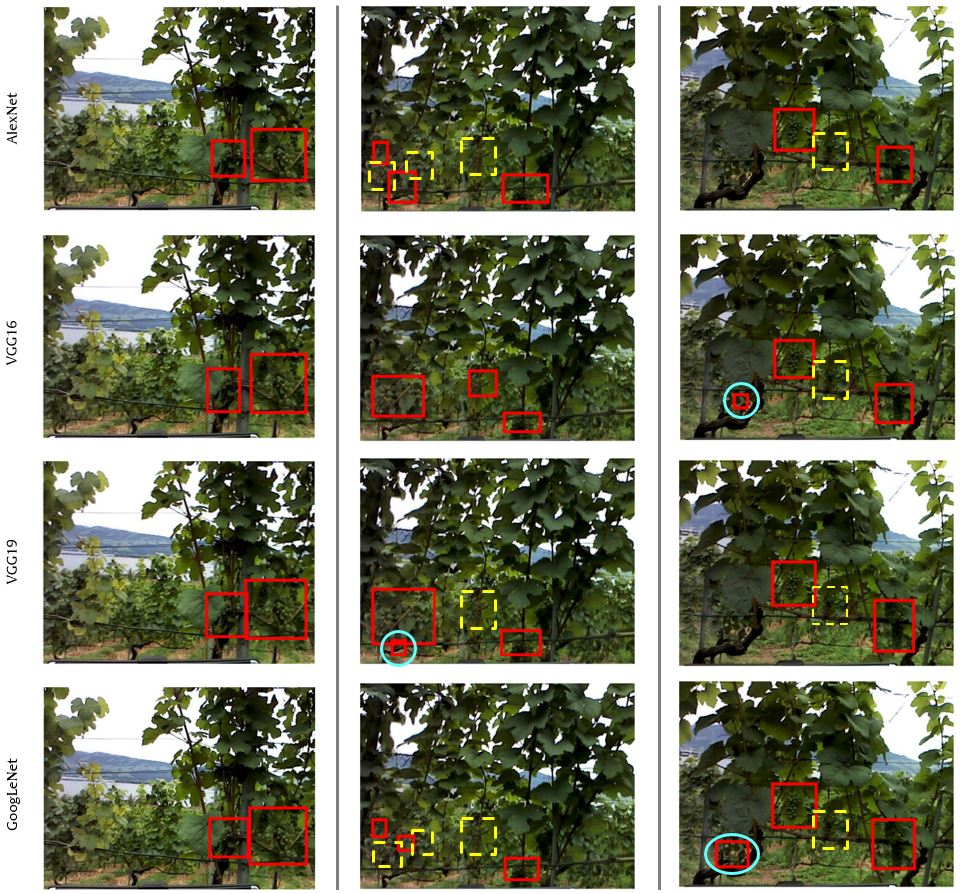}
      \caption{Sample images with bounding boxes (red rectangles) enclosing the detected grape clusters. The dashed yellow rectangles locate missed detections of grape cluster, whereas the cyan circles enclose false detections.}
      \label{fig::grape_example}
\end{figure}

At each position of the window, these scores can be rearranged to create 5 probability images. Since image analysis is aimed at the detection of grape clusters and its coarse segmentation through a bounding box, the probability map of the bunch class has to be further processed. Specifically, the probability maps of class bunch obtained by the classification of the 25 test images are converted to binary images, based on a threshold, whose level is fixed to 0.85. As a consequence, all bunch detections performed with scores higher than 85\% are grouped in connected components. These regions are also processed by simple morphological filters: a morphological closing (dilation followed by erosion) on the binary image is performed using a $5\times5$ structural element of circular shape. Finally, the smallest rectangles enclosing the resulting connected components of the binary images determine the bounding boxes of the detected grape clusters. Three representative images, output of each CNN, are presented in Figure \ref{fig::grape_example}, together with the bounding boxes (red rectangles) enclosing the detected grape clusters.

Bounding boxes in Figure \ref{fig::grape_example} well include grape clusters, although their guess suffers from the overestimation of the extension of grape clusters. It is due to the use of the moving window, since patches with grape cluster placed at their edges are entirely labeled as bunches. In addition, this overestimation of bunch sizes also results from the specific choice of the threshold level (0.85 in these experiments). This value has been set to obtain a good balance between overestimation of grape regions, missed detections (yellow rectangles in Figure \ref{fig::grape_example}), and false positives (cyan circles of Figure \ref{fig::grape_example}). For instance, if the threshold is increased, overestimation and false positives can vanish, but at the same time further missed detections can arise. With this setting of the threshold value, only few small grape clusters are not detected as effect of the low quality of the input image in terms of low resolution and significant motion blur. At the same time only small regions of actual ambiguity are marked as bunches. As an example, the cyan circles of Figure \ref{fig::grape_example} resulting from the VGG16 and the GoogLeNet are the only false detections obtained by running the tests on the whole dataset. Quantitative results in terms of metrics from Eq. \ref{eq7} to \ref{eq9} are finally reported in Table \ref{tab::test_results} for each proposed CNN.

\begin{table}[]
  \centering
  \caption{Results of tests on 25 input images, computed accordingly with metrics from Eq. \ref{eq7} to \ref{eq9}.}
  \label{tab::test_results}
  \begin{tabular}{lccccc}
      & $F_{GC}/GC$ & $N_{GC}/GC$ & $P_{GC}$ & $R_{GC}$  & $Acc_{GC}$\\
      CNN & (\%) & (\%) & (\%) & (\%) & (\%)\\
     \hline
    AlexNet & 12.50 	&14.30	&87.03	&85.45	&81.03\\
    VGG16  & 1.90	&17.30	&98.0	&84.48	&83.05\\
    VGG19  & 12.30	&10.70	&87.09	&88.52	&91.52\\
    GoogLeNet & 2.10 	&29.16	&97.91	&77.04	&79.66
  \end{tabular}
\end{table}

Results in Table \ref{tab::test_results} demonstrate that all CNNs can effectively count fruits regardless the low quality of the input color images. A deeper insight shows that the VGG19 returns the best accuracy (91.52\%) and recall (88.52\%) with respect to the other CNNs, although its precision (87.09\%) is worse than the ones provided by the GoogLeNet (97.91\%) and the VGG16 (98.0\%). In particular, the latter CNN can significantly reduce the number of false predictions to values less than 2\% (see the cyan circles of Figure \ref{fig::grape_example}). This means that all recognitions of grape clusters effectively refer to actual fruits. At the same time the VGG16 produce lower false negatives, i.e. missed recognitions, than the GoogLeNet, suggesting that it is more sensitive to the presence of grape clusters with respect to the GoogLeNet. However, the AlexNet shows balanced results in terms of precision and recall, which are almost comparable. Although results from the AlexNet are not the best of this comparison, it can be preferable because of its lower requirements in terms of processing time.


\section{Conclusions}\label{s5}
This paper proposed an in-field high throughput grapevine phenotyping platform using an Intel RealSense R200 depth camera mounted on-board an agricultural vehicle. The consumer-grade sensor can be a rich source of information from which one can infer important characteristics of the grapevine during vehicle operations. Two problems were addressed: canopy volume estimation and grape bunch detection. \\Different modeling approaches were evaluated for plant per plant volume estimation starting from a 3D point cloud of a grapevine row and they were compared with the results of a manual measurement procedure. It was shown that depending on the adopted modeling approach different results can be obtained. That indicates the necessity of establishing new standards and procedures. Four deep learning frameworks were also implemented to segment visual images acquired by the RGB-D sensor into multiple classes and recognize grape bunches. Despite the low quality of the input images, all methods were able to correctly detect fruits, with a maximum value of accuracy of 91.52\%, obtained by the VGG19. Overall, it was shown that the proposed framework built upon sensory data acquired by the Intel RealSense R200 is effective in measuring important characteristics of grapevine in the field in a non-invasive and automatic way.

\section*{\emph{Future work}}

Both perception methods discussed in the paper use a consumer-grade camera (worth a few hundred Euros) as the only sensory input. This choice proved to be a good trade-off between performance and cost-effectiveness. An obvious improvement would be the adoption of higher-end depth cameras available on the market. A new model of RealSense (D435 Camera) was just delivered by Intel featuring improved resolution and outdoor performance. Another common issue was the large vibrations experienced by the camera during motion that was induced by the on-board combustion engine and the terrain irregularity. This calls for a specific study to model the vehicle response and compensate/correct for this disturbance source.
The first method for automatic volume estimation comprises three main stages: 3D grapevine row reconstruction, canopy segmentation, and single plant volume estimation. Each stage may be refined improving the overall performance. For example, the 3D reconstruction of the entire row strongly depends on the pose estimation of the operating vehicle. In the proposed embodiment, the robot localization system uses only visual information. However, visual odometry typically suffer from drift or accumulation errors, especially for long range and long duration tasks. Therefore, the integration of complementary sensor types, including moderate-cost GPS and IMU (e.g., \citep{FLE}), may significantly enhance this stage. In addition, the reliance on simultaneous localization and mapping (SLAM) approaches may be certainly beneficial and it would allow continuous estimation between successive crop rows mitigating the difficulties connected with the 180-deg steering stage at the end of each row.
The canopy segmentation step was performed using fixed threshold parameters that were set empirically to give the best results. However, the adoption of adaptive thresholds may lead to improved canopy segmentation. In this respect, the integration of ground classifiers (e.g., \citep{MIL1}) may help to better extract the canopy from the rest of the scene. Finally, the presence of voids and concavities play a significant role in the accuracy of the canopy volume estimation. Global methods that use the entire 3D point cloud (e.g., CH, OBB, and AABB) tend to provide higher estimates, whereas local approaches (e.g., 3D-OG) may fit better the canopy volume. Other alternatives could be used, such as segmented convex hull and alpha shape. However, it should be noted that voids in 3D reconstruction may also be due to occlusions or failures in the stereo processing. Hence, methods that seemingly fit better the point cloud may not necessarily lead to more accurate estimates of the real plant volume.
\\Further work will also regard the topic of grape cluster detection. As a first step, next investigations will include the optimization of the whole processing on a dedicated hardware, in order to reduce processing time, till reaching real-time computations. In addition, the same deep-learning approaches will be extended to the recognition of further fruits, such as green or red grapes, or at different days during fruits maturation. For this reason, extended data sets must be considered to improve the learning phase of the CNNs, aimed at labeling input images on a higher number of classes, which will depend on age, kind and color of the fruits. Finally, future investigations will be also dedicated to the use of 3D information to improve classification results. First, 3D information will be managed to determine the best patch size, now fixed at a size of 80$\times$80 pixels. Exploiting 3D information, the choice of the patch size will be adjusted dynamically on the actual plant volume and on the distance between the plant and the sensor. At the same time, a completely different approach will be pursued: depth maps, aligned on the corresponding color images, will constitute the fourth channel of the input patches. In this case, the complete architecture of the CNNs will be redefined, and thus trained from the scratch, to receive data of different dimension.

\section*{Acknowledgments}

The financial support of the following grants are acknowledged: Simultaneous Safety and Surveying for Collaborative Agricultural Vehicles (S3-CAV), ERA-NET ICT-AGRI2 (Grant No. 29839), Autonomous DEcision Making in very long traverses (ADE), H2020 (Grant No. 821988), and Electronic Shopping $\&$ Home delivery of Edible goods with Low environmental Footprint (E-SHELF), POR Puglia FESR-FSE 2014-2020(Id. OSW3NO1). The authors are also grateful to Stefan Rilling, Peter Fr\"{o}lich and Michael Nielsen for the valuable support in performing the experiments and gathering the sensory data.
\bibliography{CEA_GR}

\end{document}